\documentclass[cameraready]{Interspeech}

\title{Reinforcement Learning for Data-Efficient Code-Switched ASR}

\author[affiliation={1}, orcid=0009-0006-6847-7368, correspondingauthor]{Ziwei}{Ye}
\author[affiliation={2}, orcid=0000-0003-0479-7363]{Peter}{Vickers}

\address{
    $^1$ Independent Researcher\\
    $^2$ Spotify Canada
}

\email{zxy1677@rit.edu, pvickers@spotify.com}


\usepackage{comment}
\usepackage{amsmath}
\usepackage{stfloats}
\usepackage[LAE,T2A,T1]{fontenc}
\usepackage[utf8]{inputenc}
\usepackage{CJKutf8}
\usepackage{multirow}
\usepackage{pgfplots}
\pgfplotsset{compat=1.18}
\newcommand{\textcjk}[1]{\begin{CJK*}{UTF8}{gbsn}#1\end{CJK*}}

\begin{document}

\maketitle

\begin{abstract}
Audio-language models can be prompted for code-switched speech, but their decoding is not optimized for code-switching and often fails at language boundaries. We propose a practical reinforcement learning with verifiable rewards recipe for data-efficient adaptation of audio-language models to code-switched ASR using group relative policy optimization, combining an error rate reward with a script fidelity reward that penalizes wrong writing systems and a two-pass draft-and-refinement procedure. Using Qwen2-Audio as a reproducible testbed across 10 language pairs, training on only TTS code-switched speech, we show that RLVR with 10\% of the data matches LoRA supervised fine-tuning trained on the full dataset, with the largest gains on typologically distant pairs. The error rate reward eliminates translation errors while the script fidelity reward separately reduces script contamination without degradation. These gains transfer zero-shot to a human-recorded code-switching corpus.
\end{abstract}

\section{Introduction}
Code-switching is a natural communicative behavior for the majority of the world's multilingual population \cite{poplack1980codeswitching, dogruoz2021survey}.
Despite its prevalence, code-switched automatic speech recognition (CS-ASR) remains challenging due to the need to model rapidly shifting acoustic and linguistic cues, compounded by the scarcity of labeled code-switched training data \cite{winata2023decades, sitaram2019survey}.
Since large-scale speech models are typically pre-trained on monolingual corpora \cite{radford2022whisper, baevski2020wav2vec, chu2023qwenaudio}, data-efficient adaptation methods are needed to endow speech-augmented large language models (speech-LLMs) with robustness to code-switching.

Instruction-following speech-LLMs can ingest language context via prompting for flexible output control, yet their autoregressive decoding is trained with token-level cross-entropy, which does not directly optimize sequence-level metrics such as CER and is subject to exposure bias \cite{ranzato2016sequence}.
These issues can be amplified at code-switch boundaries, where language confusion and script hallucination have been observed \cite{liu2023language_confusion}.

We frame CS-ASR as a verifiable reward problem: given reference transcripts,
we can compute automatic rewards
and use RL to directly optimize sequence-level transcription quality.
Rather than targeting state-of-the-art performance, we use an established, publicly available speech-LLM (Qwen2-Audio) as a controlled testbed to study how reward design and data efficiency interact for code-switched adaptation. In this paper, we make the following contributions:

\begin{itemize}
  \item We apply \textbf{Reinforcement Learning with Verifiable Rewards (RLVR)} \cite{shao2024deepseekmath} on code-switched ASR, using rewards based on CER and a \textbf{script fidelity reward} to directly encourage correct writing systems at code-switch boundaries.
  \item We introduce a \textbf{training-time two-pass draft and refinement} procedure that conditions a second decoding pass
  on the best draft to encourage ``listen again and fix'' behavior.
  \item We evaluate on the human-read subset of CS-FLEURS \cite{yan2025csfleurs} across 10 language pairs, showing data-efficient gains over LoRA supervised fine-tuning (SFT), and demonstrate zero-shot transfer to SwitchLingua \cite{xie2025switchlingua}, a human-recorded code-switching corpus.
\end{itemize}

\definecolor{errtransl}{RGB}{200,40,40}
\definecolor{errgarble}{RGB}{180,100,0}
\definecolor{rewcer}{RGB}{30,120,180}
\definecolor{rewrefine}{RGB}{100,60,160}
\definecolor{correct}{RGB}{30,130,60}

\begin{figure}[t]
  \centering
  \small
  \setlength{\fboxsep}{1.5pt}%
  \newcommand{\errhl}[1]{\colorbox{errtransl!15}{#1}}%
  \newcommand{\garhl}[1]{\colorbox{errgarble!15}{#1}}%
  \begin{tabular}{@{}p{\columnwidth}@{}}
    \toprule
    \textbf{Reference} (cmn-eng, SwitchLingua)\\
    \textcjk{这种科学发现真是太}fascinating\textcjk{了}\\[3pt]
    \midrule
    \textbf{Format-only baseline} \hfill \textcolor{errtransl}{CER 1.81}\\
    \errhl{\textit{This kind of scientific discovery is really}} \garhl{\textit{fascinating}}\\
    {\scriptsize
      \textcolor{errtransl}{\errhl{Chinese translated to English}} \quad
      \textcolor{errgarble}{\garhl{\textcjk{太}...fascinating...\textcjk{了} frame code-switching lost}}}\\[3pt]
    \midrule
    \textbf{LoRA SFT} (100\% data) \hfill \textcolor{errtransl}{CER 1.33}\\
    \errhl{\textit{This scientific discovery is so}} \garhl{\textit{fascinating.}}\\
    \midrule
    \textbf{RLVR} (10\% data) \hfill \textcolor{correct}{CER 0.00}\\
    \textcjk{这种科学发现真是太}fascinating\textcjk{了}\\
    {\scriptsize
      \textcolor{correct}{~Chinese preserved} \quad
      \textcolor{correct}{~English insertion intact} \quad
      \textcolor{correct}{~Code-switch boundary correct}}\\[3pt]
    \bottomrule
  \end{tabular}
  \caption{Chinese-English code-switch sample in SwitchLingua}
  \label{fig:example}
\end{figure}

\section{Related work}

\noindent\textbf{Code-switched ASR.}
Code-switching ASR must jointly model acoustics and rapidly shifting language priors, and is challenged by language confusion errors and limited labeled code-switched data \cite{mustafa2022codeswitchreview}.
Recent work incorporates explicit language tag signals and language-aware decoding \cite{wang2023textderivedlid}.
FLEURS provides a standardized multilingual benchmark \cite{conneau2022fleurs}, and CS-FLEURS extends it to code-switched pairs \cite{yan2025csfleurs}.

\noindent\textbf{Speech-LLMs for ASR.}
Large-scale speech encoder-decoder models such as Whisper \cite{radford2022whisper} have motivated speech-language models that pair speech encoders with LLM decoders for flexible instruction following \cite{chu2023qwenaudio, lai2023instructionfollowing}.
A recurring challenge is hallucination, and in neural machine translation, exposure bias has been linked to hallucination under domain shift \cite{wang2020exposure_bias}, an analogy that motivates RL-based training for code-switching, where the language distribution shifts abruptly at boundaries.

\noindent\textbf{Sequence-level objectives and RLVR.}
Directly optimizing ASR evaluation metrics has a long history, including minimum word error rate training and minimum Bayes risk \cite{prabhavalkar2018mwer, shannon2017embr}.
Reinforcement learning with verifiable rewards (RLVR) extends this idea using reward signals in domains where correctness can be automatically checked \cite{wen2025rlvr}.
Group Relative Policy Optimization (GRPO) normalizes rewards within groups of sampled completions to avoid an explicit learned critic \cite{shao2024deepseekmath}, and has been used in reasoning RL \cite{guo2025deepseekr1}.
Concurrent work has applied GRPO to monolingual ASR \cite{shivakumar2025grpo_asr} and to audio question-answering \cite{rouditchenko2025omnir1}.
Our approach adapts RLVR to code-switched ASR with transcription rewards augmented by code-switch consistency signals.

\section{Method}

\subsection{Problem Setup}
Given an audio segment $\mathbf{x}$, a language context $\mathcal{C}$, and a reference transcript $y^*$, the model produces a transcript $\hat{y} \sim \pi_\theta(\cdot \mid \mathbf{x}, \mathcal{C})$. Our goal is to improve both transcription accuracy and code-switch fidelity. The language context $\mathcal{C}$ consists of a instruction specifying the language pair (``The audio contains speech mixing Japanese and English'') and format constraints.
\subsection{RLVR with GRPO}
We train the model with GRPO, a RL method that learns
from relative preferences among a small set of sampled candidate transcriptions.

For each input $(\mathbf{x}, \mathcal{C})$, we sample $G$ candidate transcriptions
$\{\hat{y}_i\}_{i=1}^{G}$ from the current policy $\pi_{\theta_{\mathrm{old}}}$ and score each one
with a verifiable reward $R_i = R(\hat{y}_i, y^*, \mathcal{C})$. We then z-score rewards
within the group to obtain advantages $A_i$, so candidates that score above the group mean receive
positive advantage and those below receive negative advantage. This encourages the model to increase
the probability of higher scoring candidates relative to their peers, without requiring a learned
value function.
  
We update the policy using a clipped surrogate objective with KL regularization to prevent
distribution collapse:
\begin{equation}
\begin{split}
\mathcal{L}(\theta) = - \mathbb{E}_{i} \Big[ \min \Big(\rho_i A_i,  \text{clip}&(\rho_i, 1-\varepsilon, 1+\varepsilon) A_i \Big) \Big] \\
& + \lambda_{\text{KL}} \text{KL}(\pi_\theta \parallel \pi_{\text{ref}})
\end{split}
\end{equation}

where
\begin{equation}
\rho_i = \frac{\pi_\theta(\hat{y}_i \mid \mathbf{x}, \mathcal{C})}{\pi_{\theta_{\text{old}}}(\hat{y}_i \mid \mathbf{x}, \mathcal{C})}
\end{equation}
is the importance ratio between the updated policy and the sampling policy.

\subsection{Script Fidelity (SHR) Reward}

When the base model is tuned for structured output, it exhibits pervasive script contamination: the output contains characters from a writing system unrelated to either language in the pair.
On SwitchLingua, 18\% of non-Latin-script samples contain wrong-script characters, rising to 27\% for Japanese and 27\% for Arabic.
To discourage this, we augment the CER reward with a binary bonus for clean outputs:

$R(\hat{y}_i, y^*, \mathcal{C}) = -\mathrm{CER}(\hat{y}_i,\, y^*) + \beta_{\mathrm{sf}} \cdot \mathrm{Script}(\hat{y}_i, \mathcal{C})$,
where $\mathrm{Script}(\hat{y}_i, \mathcal{C}) = 1$ if all characters in $\hat{y}_i$ belong to the allowed scripts and $0$ otherwise.
The set of allowed scripts is $\mathcal{S} = \mathcal{S}_{L_1} \cup \mathcal{S}_{L_2}$, the union of Unicode scripts used by both languages in the pair. Non-alphanumeric characters such as punctuation, digits, and whitespace are always permitted. We set $\beta_{\mathrm{sf}} = 0.05$. Because GRPO normalizes rewards within each group, this bonus only affects the advantage when generations differ in script contamination.

\subsection{Two-Step Draft Refinement}
\label{sec:twostep}

Beyond script contamination, the base model reveals further failures,
On SwitchLingua, 37\% of non-Latin-script samples are fully translated into English, and 36\% of code-switched utterances retain only the English portion.
These errors suggest that single pass decoding can commit early to an incorrect language mode.
Inspired by iterative self-refinement in LLMs \cite{madaan2023selfrefine}, we address this with a two-step procedure in which the first pass generates $G$ drafts and applies a GRPO update, then refines the highest reward draft in a second GRPO pass conditioned on the original audio and the draft (Algorithm~\ref{alg:twostep}).

This two-step approach trains the model to both transcribe and self-correct within a single training iteration, without requiring any external correction signal beyond the existing reward function. We do single pass inference as with the other models during testing time.

\begin{algorithm}[t]
\caption{Two-Pass GRPO with Draft Refinement}
\label{alg:twostep}
\begin{algorithmic}[1]
\Require Audio $\mathbf{x}$, reference $y^*$, language context $\mathcal{C}$, number of generations $G$
\Statex
\Statex \textbf{Pass 1 (Draft):}
\State Sample $G$ drafts: $\hat{y}^{(g)} \sim \pi_\theta(\cdot \mid \mathbf{x}, \mathcal{C})$ for $g = 1, \ldots, G$
\State Compute rewards $R^{(g)} = R(\hat{y}^{(g)}, y^*, \mathcal{C})$ and GRPO loss $\mathcal{L}_1$
\State Select best draft: $\hat{y}^{\dagger} = \hat{y}^{(\arg\max_g R^{(g)})}$
\State Update $\theta$ with $\nabla_\theta \mathcal{L}_1$
\Statex
\Statex \textbf{Pass 2 (Refine):}
\State Construct prompt: $(\mathbf{x}, \hat{y}^{\dagger}, \mathcal{C})$
\State Sample $G$ refinements: $\tilde{y}^{(g)} \sim \pi_\theta(\cdot \mid \mathbf{x}, \hat{y}^{\dagger}, \mathcal{C})$
\State Compute rewards $\tilde{R}^{(g)} = R(\tilde{y}^{(g)}, y^*, \mathcal{C})$ and GRPO loss $\mathcal{L}_2$
\State Update $\theta$ with $\nabla_\theta \mathcal{L}_2$
\end{algorithmic}
\end{algorithm}

\begin{table*}[t]
  \caption{CER on CS-FLEURS \texttt{read\_test} and SwitchLingua (zero-shot transfer). For each data setting, the best per language pair is in \textbf{bold}. Multi-seed columns ($n$=3) report mean{\tiny$\pm$std\,(SHR)}. All models are trained on CS-FLEURS \textsc{XTTS-Train}.}
  \label{tab:results}
  \centering
  \footnotesize
  \begin{tabular}{lccccccc}
    \toprule
    & \multicolumn{3}{c}{\textbf{10\% data ($n$=2,310)}} & \multicolumn{2}{c}{\textbf{20\% data ($n$=4,625)}} & \textbf{100\% data} & \\
    \cmidrule(lr){2-4} \cmidrule(lr){5-6} \cmidrule(lr){7-7}
    \textbf{Lang pair} &
    \textbf{$\mathrm{CER}$+$\mathrm{SHR}$+refine} &
    \textbf{$\mathrm{CER}$+refine} &
    \textbf{$\mathrm{CER}$ only} &
    \textbf{$\mathrm{CER}$+$\mathrm{SHR}$+refine} &
    \textbf{LoRA SFT} &
    \textbf{LoRA SFT} &
    \textbf{Base} \\
    \midrule
    \multicolumn{8}{l}{\textit{CS-FLEURS \texttt{read\_test}}} \\
    \midrule
    deu-eng  & \textbf{.102}{\tiny$\pm$.008\,(.000)} & .112{\tiny$\pm$.001\,(.007)} & .115{\tiny$\pm$.001\,(.002)} & \textbf{.107}{\tiny$\pm$.001\,(.002)} & .123{\tiny$\pm$.002\,(.008)} & .079{\tiny$\pm$.003\,(.003)} & .288 \\
    fra-eng  & \textbf{.087}{\tiny$\pm$.002\,(.003)} & .089{\tiny$\pm$.001\,(.002)} & .093{\tiny$\pm$.001\,(.000)} & \textbf{.084}{\tiny$\pm$.000\,(.000)} & .129{\tiny$\pm$.005\,(.011)} & .075{\tiny$\pm$.001\,(.002)} & .297 \\
    ita-eng  & \textbf{.076}{\tiny$\pm$.003\,(.000)} & .080{\tiny$\pm$.001\,(.003)} & .083{\tiny$\pm$.000\,(.003)} & \textbf{.078}{\tiny$\pm$.002\,(.000)} & .103{\tiny$\pm$.001\,(.009)} & .067{\tiny$\pm$.004\,(.003)} & .251 \\
    por-eng  & \textbf{.086}{\tiny$\pm$.004\,(.001)} & .095{\tiny$\pm$.002\,(.003)} & .097{\tiny$\pm$.002\,(.003)} & \textbf{.088}{\tiny$\pm$.001\,(.004)} & .137{\tiny$\pm$.007\,(.004)} & .069{\tiny$\pm$.001\,(.007)} & .284 \\
    spa-eng  & \textbf{.055}{\tiny$\pm$.002\,(.002)} & .061{\tiny$\pm$.001\,(.000)} & .061{\tiny$\pm$.000\,(.002)} & \textbf{.055}{\tiny$\pm$.001\,(.002)} & .087{\tiny$\pm$.003\,(.006)} & .041{\tiny$\pm$.001\,(.003)} & .181 \\
    \cmidrule{1-8}
    ara-eng  & .269{\tiny$\pm$.003\,(.080)} & \textbf{.267}{\tiny$\pm$.004\,(.143)} & .275{\tiny$\pm$.002\,(.228)} & \textbf{.254}{\tiny$\pm$.003\,(.041)} & .531{\tiny$\pm$.004\,(.233)} & .338{\tiny$\pm$.009\,(.177)} & .883 \\
    cmn-eng  & .112{\tiny$\pm$.002\,(.000)} & \textbf{.111}{\tiny$\pm$.002\,(.001)} & .119{\tiny$\pm$.001\,(.002)} & \textbf{.102}{\tiny$\pm$.002\,(.000)} & .266{\tiny$\pm$.011\,(.002)} & .097{\tiny$\pm$.008\,(.003)} & .579 \\
    jpn-eng  & \textbf{.185}{\tiny$\pm$.002\,(.008)} & .191{\tiny$\pm$.006\,(.023)} & .210{\tiny$\pm$.011\,(.038)} & \textbf{.183}{\tiny$\pm$.001\,(.003)} & .392{\tiny$\pm$.005\,(.003)} & .230{\tiny$\pm$.009\,(.003)} & .723 \\
    kor-eng  & .229{\tiny$\pm$.003\,(.038)} & \textbf{.222}{\tiny$\pm$.003\,(.107)} & .229{\tiny$\pm$.000\,(.111)} & \textbf{.212}{\tiny$\pm$.003\,(.028)} & .359{\tiny$\pm$.028\,(.142)} & .187{\tiny$\pm$.001\,(.061)} & .635 \\
    rus-eng  & .182{\tiny$\pm$.005\,(.058)} & \textbf{.177}{\tiny$\pm$.001\,(.076)} & .182{\tiny$\pm$.005\,(.096)} & \textbf{.174}{\tiny$\pm$.002\,(.040)} & .327{\tiny$\pm$.006\,(.125)} & .196{\tiny$\pm$.012\,(.050)} & .528 \\
    \cmidrule{1-8}
    Avg CER (ma/mi) & \textbf{.138/.155} & .141/.155 & .146/.162 & \textbf{.134/.147} & .245/.295 & .138/.159 & .465/.553 \\
    Average\ SHR    & \textbf{.025}{\tiny$\pm$.003} & .048{\tiny$\pm$.001} & .068{\tiny$\pm$.005} & \textbf{.015}{\tiny$\pm$.000} & .074{\tiny$\pm$.000} & .049{\tiny$\pm$.003} & .104 \\
    \midrule
    \multicolumn{8}{l}{\textit{SwitchLingua (zero-shot transfer)}} \\
    \midrule
    deu-eng  & \textbf{.194}{\tiny$\pm$.000\,(.016)} & .197{\tiny$\pm$.005\,(.011)} & .197{\tiny$\pm$.000\,(.015)} & \textbf{.192}{\tiny$\pm$.000\,(.008)} & .211{\tiny$\pm$.002\,(.020)} & .152{\tiny$\pm$.006\,(.011)} & .324 \\
    fra-eng  & \textbf{.131}{\tiny$\pm$.002\,(.012)} & .139{\tiny$\pm$.006\,(.010)} & .133{\tiny$\pm$.001\,(.012)} & \textbf{.130}{\tiny$\pm$.000\,(.005)} & .172{\tiny$\pm$.003\,(.014)} & .120{\tiny$\pm$.001\,(.006)} & .276 \\
    ita-eng  & \textbf{.120}{\tiny$\pm$.002\,(.022)} & .127{\tiny$\pm$.010\,(.017)} & .129{\tiny$\pm$.000\,(.027)} & \textbf{.120}{\tiny$\pm$.000\,(.009)} & .198{\tiny$\pm$.003\,(.031)} & .103{\tiny$\pm$.011\,(.020)} & .329 \\
    \cmidrule{1-8}
    ara-eng  & \textbf{.408}{\tiny$\pm$.013\,(.084)} & .410{\tiny$\pm$.013\,(.141)} & .411{\tiny$\pm$.007\,(.210)} & \textbf{.393}{\tiny$\pm$.005\,(.044)} & .831{\tiny$\pm$.002\,(.192)} & .652{\tiny$\pm$.005\,(.149)} & 1.031 \\
    cmn-eng  & .109{\tiny$\pm$.002\,(.000)} & .102{\tiny$\pm$.023\,(.000)} & \textbf{.101}{\tiny$\pm$.004\,(.000)} & \textbf{.089}{\tiny$\pm$.010\,(.000)} & .348{\tiny$\pm$.016\,(.000)} & .076{\tiny$\pm$.001\,(.000)} & .654 \\
    jpn-eng  & .192{\tiny$\pm$.008\,(.004)} & \textbf{.189}{\tiny$\pm$.020\,(.007)} & .193{\tiny$\pm$.003\,(.016)} & \textbf{.177}{\tiny$\pm$.008\,(.003)} & .383{\tiny$\pm$.013\,(.005)} & .188{\tiny$\pm$.012\,(.000)} & .539 \\
    kor-eng  & \textbf{.473}{\tiny$\pm$.013\,(.318)} & .509{\tiny$\pm$.005\,(.567)} & .493{\tiny$\pm$.017\,(.561)} & \textbf{.503}{\tiny$\pm$.010\,(.312)} & .975{\tiny$\pm$.047\,(.202)} & .536{\tiny$\pm$.079\,(.125)} & 1.001 \\
    rus-eng  & .242{\tiny$\pm$.006\,(.034)} & .239{\tiny$\pm$.006\,(.087)} & \textbf{.233}{\tiny$\pm$.002\,(.103)} & \textbf{.230}{\tiny$\pm$.002\,(.030)} & .437{\tiny$\pm$.013\,(.097)} & .290{\tiny$\pm$.012\,(.055)} & .541 \\
    \cmidrule{1-8}
    Avg CER (ma/mi) & \textbf{.234/.224} & .239/.229 & .236/.226 & \textbf{.229/.219} & .444/.450 & .265/.246 & .587/.609 \\
    Average SHR   & \textbf{.061}{\tiny$\pm$.011} & .105{\tiny$\pm$.002} & .114{\tiny$\pm$.008} & \textbf{.053}{\tiny$\pm$.003} & .059{\tiny$\pm$.000} & .038{\tiny$\pm$.002} & .071 \\
    \midrule
    Time/Step (8×A100)   & 1.5m & 1.5m & 1.5m & 1.5m & 3.6s & 3.6s & -- \\
    Training Steps & 132 & 132 & 132 & 264 & 528 & 5,176 & -- \\
    Total       & 204m & 204m & 202m & 396m & 38m & 315m & -- \\
    \bottomrule
  \end{tabular}
\end{table*}

\section{Experimental setup}
\subsection{Datasets}
\noindent\textbf{CS-FLEURS.}
CS-FLEURS \cite{yan2025csfleurs} is a massively multilingual code-switched speech dataset built on FLEURS \cite{conneau2022fleurs}, covering 52 languages across 113 code-switched pairs. Code-switched text is generated with an align-then-swap procedure that replaces 30\% of content words using word alignments from AwesomeAlign \cite{dou2021word}, yielding consistent code-mixing across pairs. We use two of its splits: \textsc{XTTS-Train} (128\,h, 16 X-English pairs), where speech is synthesized with XTTS-v2 \cite{casanova2024xtts}, for training, and \textsc{Read-Test} (3.3\,h, 14 X-English pairs) which contains human-recorded bilingual speech for evaluation. Text includes English markers denoting code-switched spans and our normalization strips Unicode punctuation.

\noindent\textbf{SwitchLingua.}
We additionally evaluate on SwitchLingua \cite{xie2025switchlingua}, a multilingual code-switching dataset with over 80 hours of human-recorded audio across 12 X-English pairs in single- and multi-turn dialogue formats. We restrict evaluation to the 8 pairs supported by Qwen2-Audio (ara, cmn, deu, fra, ita, jpn, kor, rus), excluding Cantonese, Hindi (unsupported by the base model) and Spanish (mislabeled in the Huggingface dataset). SwitchLingua complements CS-FLEURS by providing naturally structured code-switching with human-recorded audio rather than TTS-synthesized speech.

\subsection{Baselines}
We compare against two baselines that isolate the contribution of RLVR:
\begin{itemize}
  \item \textbf{Prompt-only (format)}: A base Qwen2-Audio model trained for a small number of iterations with a format-only reward that encourages outputting transcripts inside \texttt{<answer>} tags, without any transcription-quality signal.
  \item \textbf{LoRA SFT}: LoRA \cite{hu2022lora} fine-tuning on the full CS-FLEURS training set, providing an upper bound on what supervised training alone can achieve at this scale.
\end{itemize}

\subsection{Training details}
\noindent\textbf{Base model.}
All experiments start from Qwen2-Audio-7B-Instruct \cite{chu2023qwenaudio}, a 7B-parameter audio-language model with a Whisper-based audio encoder \cite{radford2022whisper} and a Qwen-2 LLM decoder \cite{yang2024qwen2}, trained in \texttt{bfloat16}. We choose Qwen2-Audio as a controlled, publicly available testbed: its open weights enable full reproducibility, and its weaknesses on code-switched speech provide a starting point for studying reward design. Because Qwen2-Audio does not cover all CS-FLEURS language pairs, we restrict experiments to the 10 pairs where the base model has non-trivial recognition capability.

\noindent\textbf{GRPO (RLVR).}
Both RLVR and LoRA keep the audio encoder frozen and update only the LLM decoder. RLVR updates all decoder parameters, while LoRA updates low-rank adapters. We train with GRPO using DeepSpeed ZeRO-2 \cite{rasley2020deepspeed} on 8 GPUs (clip $\varepsilon{=}0.2$, initial checkpoint as $\pi_{\text{ref}}$). We sample $G{=}8$ completions per prompt at temperature $\tau{=}1.0$, max length 512, with per-device batch size 1, gradient accumulation 8, and a constant learning rate of $10^{-6}$. Two-pass models train for 4 epochs (2 updates per step). Single-pass models train for 8 epochs, equalizing gradient updates. All evaluation uses single-pass greedy decoding. A format reward requiring \texttt{<answer>} tags is shared across all configurations. We subsample 2\,310 utterances (10\%) or 4\,625 (20\%) from \textsc{XTTS-Train} across 10 language pairs. Both RLVR and LoRA experiments are run with 3 identical random seeds.

\noindent\textbf{LoRA SFT.}
The LoRA baseline \cite{hu2022lora} applies low-rank adapters ($r{=}64$, $\alpha{=}128$, dropout $0.05$) to all attention and MLP projections of the LLM decoder. Training matches GRPO in batch size, accumulation, and learning rate, with 4 epochs on the full \textsc{XTTS-Train} split ($10{\times}$ the 10\% RLVR data).

\subsection{Metrics}
\textbf{Character Error Rate (CER)}: the Levenshtein edit distance between the hypothesis and reference sequences, divided by the reference length. Both strings are lowercased, stripped of punctuation, and have whitespace removed before comparison: $\mathrm{CER} = \mathrm{Lev}(\hat{y},\, y^*) \,/\, |y^*|$.
We report both macro-averaged CER over language pairs and micro-averaged CER, as the two can diverge when pairs differ in reference length.

\noindent\textbf{Script Hallucination Rate (SHR)}: the fraction of samples containing any character from a disallowed Unicode script. \ $\mathrm{SHR} = \frac{1}{N}\sum_{n} \mathbf{1}[\exists\, c \in \hat{y}_n : \mathrm{script}(c) \notin \mathcal{S}]$. This is the evaluation-time counterpart of the SHR reward (Section~3.3).

\section{Results}
  \noindent\textbf{CS-FLEURS human-read speech.}
  Table~\ref{tab:results} shows that with only 20\% of the
  training data, RLVR with two-pass refinement and the SHR
  reward achieves a micro-averaged CER of $0.147$,
  outperforming 100\% LoRA SFT ($0.159$) despite using
  $5{\times}$ less data.
  Even at 10\% data, RLVR matches or surpasses LoRA trained on
  the full dataset.
  Two-pass refinement provides a clear gain over single-pass
  CER training at 10\% data.
  The SHR reward substantially reduces script hallucinations
  (SHR $0.025$ vs.\ $0.068$ for CER-only at 10\%) without
  degrading CER.
  LoRA's advantage is limited to European pairs (deu, fra,
  ita, por, spa), cmn-eng, and kor-eng, where CER is already
  below 10\%, while RLVR yields the largest gains on
  typologically distant pairs (ara, jpn, rus) where
  code-switch transitions are hardest.

  \noindent\textbf{SwitchLingua (Cross-dataset transfer).}
  To test whether RLVR generalizes beyond TTS-synthesized
  speech, we evaluate all models zero-shot on SwitchLingua, a
  human-recorded code-switching corpus with naturally
  structured multi-turn dialogues and diverse speaker
  backgrounds.
  Table~\ref{tab:results} shows that 20\% RLVR with the SHR
  reward ($0.219$ micro-averaged CER) again outperforms 100\%
  LoRA SFT ($0.246$) despite using $5{\times}$ less data,
  demonstrating that the gains transfer across domains.
  Unlike CS-FLEURS, two-pass refinement alone does not improve
  over single-pass CER training at 10\% data.
  However, adding the SHR reward recovers the gain and
  substantially reduces SHR ($0.061$ vs.\ $0.114$ for
  CER-only).
  The largest RLVR gains appear on ara-eng ($0.393$ vs.\ 
  $0.652$ for LoRA) and rus-eng, both typologically distant
  pairs where LoRA's translation failure mode is most
  pronounced.
  LoRA retains an advantage on European pairs (fra, deu, ita)
  and cmn-eng, consistent with the CS-FLEURS pattern.
  Overall CER are higher than on CS-FLEURS
  \textsc{read\_test}, reflecting the greater difficulty of
  SwitchLingua's long-form, multi-turn recordings.

\section{Analysis and Discussion}

\subsection{Do the rewards address code-switching failures?}

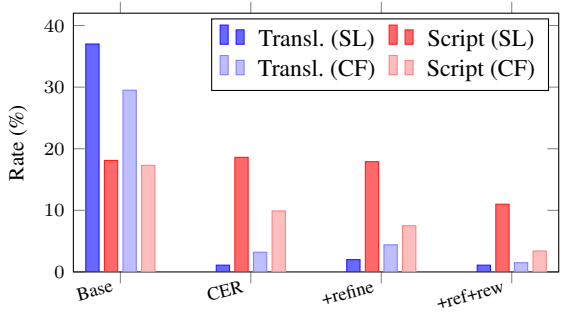
\begin{figure}[t]
  \centering
  \begin{tikzpicture}
    \begin{axis}[
      ybar,
      bar width=5pt,
      width=\columnwidth,
      height=5.0cm,
      enlarge x limits=0.12,
      ylabel={Rate (\%)},
      ylabel style={font=\footnotesize},
      symbolic x coords={Base, {CER}, {+refine}, {+ref+rew}},
      xtick=data,
      xticklabel style={font=\scriptsize, rotate=15, anchor=east},
      ymin=0, ymax=42,
      ytick={0,10,20,30,40},
      yticklabel style={font=\scriptsize},
      legend style={font=\small, at={(0.98,0.98)}, anchor=north east, legend columns=2, column sep=3pt},
      every axis plot/.append style={fill opacity=0.85},
    ]
    \addplot[fill=blue!70, draw=blue!90] coordinates {(Base,37.0) ({CER},1.1) ({+refine},2.0) ({+ref+rew},1.1)};
    \addplot[fill=red!70, draw=red!90] coordinates {(Base,18.1) ({CER},18.6) ({+refine},17.9) ({+ref+rew},11.0)};
    \addplot[fill=blue!30, draw=blue!50] coordinates {(Base,29.5) ({CER},3.2) ({+refine},4.4) ({+ref+rew},1.5)};
    \addplot[fill=red!30, draw=red!50] coordinates {(Base,17.3) ({CER},9.9) ({+refine},7.5) ({+ref+rew},3.4)};
    \legend{Transl.\ (SL), Script (SL), Transl.\ (CF), Script (CF)}
    \end{axis}
  \end{tikzpicture}
  \caption{Translation and wrong-script rates across the 10\% RLVR ablation chain on non-Latin-script samples. SL = SwitchLingua, CF = CS-FLEURS. CER reward eliminates translation. The SHR reward is needed to reduce script contamination.}
  \label{fig:failure_modes}
\end{figure}

To isolate the contribution of each reward component, we measure translation rate (non-English portion rendered as English) and wrong-script rate (characters from an unrelated writing system) across the ablation chain on non-Latin-script samples. Figure~\ref{fig:failure_modes} summarizes the results.

CER reward alone virtually eliminates translation errors (37\%$\to$1.1\% on SwitchLingua, 30\%$\to$3.2\% on CS-FLEURS), since translating produces high edit distance from the code-switched reference.
However, CER reward does not fix script contamination: wrong-script rates remain at 18.6\% on SwitchLingua with CER-only training, nearly unchanged from the base model (18.1\%).
Adding the SHR reward reduces this to 11.0\%, a 41\% relative reduction, without degrading CER.
On CS-FLEURS the effect is stronger: wrong-script rates fall from 9.9\% to 3.4\%.
For Latin-script pairs, the SHR reward also improves CER indirectly by regularizing surface-form fidelity (``seventy~km'' $\to$ ``70~km''), demonstrating the need for better evaluation metrics.

Two-pass refinement reduces wrong-script rates on CS-FLEURS (from 9.9\% to 7.5\%) even without the SHR reward, suggesting the second pass provides an opportunity to correct first-pass script errors. On SwitchLingua, refinement alone does not improve CER or wrong-script rate.
A per-pair breakdown shows refinement helps most pairs (e.g., jpn-eng $-.014$, cmn-eng $-.011$ on CS-FLEURS) but consistently hurts Arabic-English on both datasets ($+.068$ on CS-FLEURS, $+.070$ on SwitchLingua). On SwitchLingua, 5 of 8 pairs improve but the Arabic degradation tips the overall average negative.
This suggests the second pass can overcorrect on pairs where script boundaries are most ambiguous, and the SHR reward is needed to constrain such overcorrection.

\subsection{What are the cases where LoRA or RLVR fails?}

We compare 20\% RLVR (with refinement and SHR reward) against 100\% LoRA SFT.
On CS-FLEURS, RLVR outperforms LoRA by $>$0.2 CER on 275 samples, while LoRA wins on 132.
RLVR's advantage concentrates on typologically distant pairs: ara-eng (158 vs.\ 19 wins) and jpn-eng (24 vs.\ 4).
LoRA's advantage appears on kor-eng (26 vs.\ 17 wins on CS-FLEURS, 143 vs.\ 64 on SwitchLingua) and cmn-eng.

In scenarios where RLVR wins, LoRA translates code-switched spans into a single language (CER$>$0.8), while RLVR preserves code-switch boundaries (CER$<$0.15).
Conversely, RLVR garbles text or fails to preserve exact technical tokens that LoRA reproduces faithfully in cases where LoRA wins.

LoRA achieves more near-perfect transcriptions (29.3\% with CER$<$0.05 vs.\ 23.1\% for RLVR on SwitchLingua), but RLVR produces fewer catastrophic errors (10.1\% with CER$>$0.5 vs.\ 11.8\%), consistent with RL optimizing expected reward rather than memorizing individual examples.

\section{Conclusion}
We showed that RLVR with GRPO enables highly data-efficient adaptation of speech-LLMs to code-switched ASR.
With only 10\% of the training data and ${\sim}40{\times}$ fewer gradient steps, RLVR matches LoRA SFT trained on the full dataset across 10 language pairs, and at 20\% outperforms it, with the largest gains on typologically distant pairs where code-switch transitions are hardest.
A CER reward eliminates translation errors, while a dedicated SHR reward is needed to reduce script contamination, and a two-pass draft refinement paradigm has potential to mitigate both error types.
These gains transfer zero-shot from TTS-synthesized training data to human-recorded code-switching speech, demonstrating robustness across acoustic domains.

\section{Generative AI Use Disclosure}
LLMs were used to produce a fraction of the source codes for experiment orchestration. Additionally, LLMs were used for spellchecking and suggesting edits to further improve the fluency and conciseness of the paper during the drafting process.

\bibliographystyle{IEEEtran}
\bibliography{mybib}

\end{document}